\pdfoutput=1

\documentclass{article}

\usepackage[square,numbers]{natbib}

     \usepackage[nonatbib,preprint]{neurips_2021}

\usepackage[utf8]{inputenc} % allow utf-8 input
\usepackage[T1]{fontenc}    % use 8-bit T1 fonts
\usepackage{url}            % simple URL typesetting
\usepackage{booktabs}       % professional-quality tables
\usepackage{amsfonts}       % blackboard math symbols
\usepackage{nicefrac}       % compact symbols for 1/2, etc.
\usepackage{tasks}

\settasks{
  label-width = 2em ,
  item-indent = 2em ,
  before-skip = -\parskip , % undo paragraph skip
  after-skip = -\parskip , % undo paragraph skip
  after-item-skip = \parskip % undo paragraph skip
}
\usepackage{microtype}      % microtypography
\usepackage{xcolor}         % colors

\usepackage{microtype}
\usepackage{graphicx}
\usepackage{booktabs} % for professional tables
\usepackage{comment}
\usepackage{enumitem}

\usepackage{algorithm}
\usepackage[noend]{algorithmic}
\usepackage[T1]{fontenc}    % use 8-bit T1 fonts
\usepackage{url}      
\usepackage{subfiles}
\usepackage{booktabs}       % professional-quality tables
\usepackage{amsfonts}       % blackboard math symbols
\usepackage{nicefrac}  
\usepackage{xr}
\usepackage{microtype}      % microtypography
\usepackage{amsmath, mathtools, amsthm, amssymb}

\usepackage{tabularx}
\usepackage[noend]{algorithmic}

\usepackage{graphicx}
\usepackage{subfigure}

\usepackage{graphicx}
\usepackage{tabularx}
\usepackage{multicol}

\usepackage{hyperref}

\title{Policy Gradients Incorporating the Future}

\author{%
  David Venuto\textsuperscript{1,2}, Elaine Lau\textsuperscript{2}, Doina Precup\textsuperscript{1,2,3}, Ofir Nachum\textsuperscript{4} \\
  \textsuperscript{1}Mila, \textsuperscript{2}McGill University, \textsuperscript{3}DeepMind, \textsuperscript{4}Google Brain\\
  \texttt{\{david.venuto, tsoi.lau\}@mail.mcgill.ca} \\
  \texttt{dprecup@cs.mcgill.ca, ofirnachum@google.com} 
}

\begin{document}

\maketitle

\begin{abstract}
Reasoning about the future -- understanding how decisions in the present time affect outcomes in the future -- is one of the central challenges for reinforcement learning (RL), especially in highly-stochastic or partially observable environments. While predicting the future directly is hard, in this work we introduce a method that allows an agent to ''look into the future'' without explicitly predicting it. Namely, we propose to allow an agent, during its training on past experience, to observe what \emph{actually} happened in the future at that time, while enforcing an information bottleneck to avoid the agent overly relying on this privileged information. This gives our agent the opportunity to utilize rich and useful information about the future trajectory dynamics in addition to the present. Our method, Policy Gradients Incorporating the Future (PGIF), is easy to implement and versatile, being applicable to virtually any policy gradient algorithm. We apply our proposed method to a number of off-the-shelf RL algorithms and show that PGIF is able to achieve higher reward faster in a variety of online and offline RL domains, as well as sparse-reward and partially observable environments. 
\end{abstract}

\section{Introduction}

Fundamentally, reinforcement learning (RL) is composed of gathering useful information (\emph{exploration}) and assigning credit to that information (\emph{credit assignment}). Both of these problems present their own unique learning challenges. 
In this work, we focus on credit assignment, which refers to the challenge of matching observed outcomes in the future to decisions made in the past.
Humans appear to do this in a sample efficient manner \cite{Johnson-Laird18243}, and so it is natural to expect our own RL agents to do so as well.

One of the most popular approaches to credit assignment, known as \emph{model-free} RL, is to learn a value function to approximate the future return given a starting state and action. 
The value function is learned 
using experience of the agent acting in the environment via temporal difference (TD) methods~\cite{Sutton1988}, which regress the value function to a target based on a combination of groundtruth returns achieved in the environment and the approximate value function itself.
The need to \emph{bootstrap} learning of the value function on its own estimates is known to lead to difficulties in practice, where one must achieve a careful balance between bias and variance~\cite{hca,10.5555/2074022.2074088,Schulmanetal_ICLR2016,pmlr-v48-mniha16}. If a slight imbalance arises, the consequences can be disastrous for learning~\cite{10.5555/2998981.2999132,vanhasselt2018deep,10.5555/3312046}. For example, in offline RL this issue is so pronounced that algorithms must apply strong regularizations on both learned policy and value function to achieve stable performance \cite{wu2020behavior,NEURIPS2019_c2073ffa,zhang2021brac,nachum2019algaedice}.
%Beyond stability issues, in TD methods time is the only metric used to assign credit \citep{Sutton1988}; actions are assigned credit from future rewards solely based on the recency in which they occurred rather than reasoning about cause-and-effect.

The model-free approach plays dual to the \emph{model-based} approach, where an agent learns a dynamics and reward model of the environment, and then learns an agent to optimize behavior in this model. Thus, credit assignment boils down to utilizing an appropriate planning algorithm that can perform multiple rollouts in the model, effectively allowing the agent to ``look into the future'' \cite{NIPS2017_9e82757e,pascanu2017learning} to determine cause-and-effect~\cite{10.1145/122344.122377,peng-etal-2018-deeppeng-etal-2018-deep,pmlr-v119-abbas20a}. 
While model-based RL may appear more straightforward, learning an accurate model is a challenge in practice, presenting its own sample-efficiency problems \cite{wang2020benchmarking} as well as memory and computational issues \cite{Kaiser2020Model}. 
Model-based approaches are thus most beneficial when the environment exhibits some level of regularity \cite{Francois-Lavet_Bengio_Precup_Pineau_2019}.  
%The model free approach, while much easier to train, suffers from the discussed issues in efficient credit assignment due to the nature of using one step backups and value function approximation \citep{10.5555/3312046}.  These methods generally cannot reason about cause-and-effect.

%\ofir{If we mention partial obs, this is where we should do it. E.g., "Credit assignment in both model-free and model-based RL is exacerbated by partially observable environments...}
%One such challenge is approximating the values of states in problems where states are partially observable \citep{Tennenholtz_Shalit_Mannor_2020,10.5555/1622262.1622264,10.5555/2998687.2998730}. 

Beyond these issues, credit assignment in both model-free and model-based RL is further exacerbated by \emph{partially observable} environments, in which the full environment state is not known to the learning agent. Thus it is infeasible to predict future events accurately. When applied to such non-Markovian domains, model-free algorithms relying on bootstrapping and value function approximation tend to be biased~\cite{Singh94learningwithout}.  On the other hand for model-based approaches, learning an accurate dynamics model in such domains is a difficult, potentially ill-defined problem \cite{NIPS1998_6d3a1e06,NIPS2009_a2557a7b}.  

%Credit assignment is also a significant challenge in offline RL, where it manifests itself in the form of function approximators of value functions incorrectly generalizing \citep{10.5555/2998981.2999132,vanhasselt2018deep} and frequent errors that arise when these estimates are extrapolated to unseen states \citep{pmlr-v97-fujimoto19a}.
%In offline RL \citep{Lange2012}, where an agent learns from a fixed data-set, we see that online RL algorithms fail since the data-set is often poorly correlated with the distribution of the current agent policy.  It is speculated that these breakdowns are due to function approximators of value functions incorrectly generalize \citep{10.5555/2998981.2999132,vanhasselt2018deep} and frequent errors that arise when these estimates are extrapolated to unseen states \citep{pmlr-v97-fujimoto19a}. T
%These issues make credit assignment particularly difficult and lead to exploding gradients while learning them (citation needed).  This problem is especially challenging in the partially observable domain which can introduce compounded modeling errors when we estimate transition probabilities and value functions while still assuming that the data was generated from a Markovian policy \citep{fu2020d4rl,10.5555/645529.658134}. 

%\textbf{I think I would like to go back and be more concrete here.}
In this work, we aim to circumvent these challenges. 
We propose a simple modification to model-free RL that allows the learned policy and value function to ``look into the future'' but without the need to learn an accurate model.
Namely, we propose to modify the policy and value function to not only condition on the presently observed state and action but also on the subsequent trajectory (sampled by the agent as it was interacting with the environment) following this state and action. 
This way, our method mitigates potential approximation or feasibility issues in accurately modeling the future.
To ensure that the learned policy and value function remains relevant during inference (i.e., data collection) when the future trajectory is unavailable, we place an \emph{information bottleneck} \cite{michael2018on,Tishby99theinformation} on the additional inputs, encouraging the learned functions to minimize their reliance on this privileged information.
One may thus view our method as an instance of \emph{teacher forcing} or \emph{$Z$-forcing} \cite{10.5555/3295222.3295416,NIPS2016_16026d60} where our student is the learned policy and value function and the teacher is some function of the information in the future trajectory.
%Alternatively, the method can be explained as using variational inference where we derive an approximate posterior over the distribution of this downwards trajectory information.

Practically, our method, Policy Gradients Incorporating the Future (PGIF), is easy to implement. 
We use either a backwards RNN or a transformer to inject downstream information from the observed trajectories by way of latent variables, with a KL divergence regularization on these latents.  
We apply PGIF on top of a variety of off-the-shelf RL algorithms, including RNN-based PPO~\cite{schulman2017proximal}, SAC \cite{pmlr-v80-haarnoja18b}, and BRAC~\cite{wu2020behavior}, and evaluate these algorithms on online and offline RL as well as sparse-reward and partially observable environments. 
In all of these domains, we demonstrate the ability of PGIF to achieve higher returns faster compared to these existing RL algorithms on their own, thus showing that our proposed method is both versatile and beneficial in practice.

\section{Background and Notation}
We begin by providing a brief overview of the notation and preliminary concepts that we will use in our later derivations.

\textbf{Markov Decision Processes (MDPs)} MDPs are defined by a tuple $\langle \mathcal{S}, \mathcal{A}, \mathbb P, R, \rho_0, \gamma \rangle$ where $\mathcal{S}$ is a set of \emph{states}, $\mathcal{A}$ is a set of \emph{actions}, $\mathbb P$ is a transition kernel giving a probability $\mathbb P(s'|s,a)$ over next states given the current state and action, $R : \mathcal{S} \times \mathcal{A} \rightarrow [R_{\text{min}},R_{\text{max}}]$ is a reward function, $\rho_0$ is an initial state distribution, and $\gamma \in [0, 1)$ is a discount factor. 
%$s_t$ and $a_t$ are respectively the state and action of the agent at time instant $t$. 
An agent in this MDP is a stationary policy $\pi$ giving a probability $\pi(a|s)$ over actions at any state $s\in \mathcal{S}$.
A policy $\pi$ interacts with the MDP by starting at $s_0\sim\rho_0$ and then at time $t\ge0$ sampling an action $a_t\sim\pi(s_t)$ at which point the MDP provides an immediate reward $R(s_t,a_t)$ and transitions to a next state $s_{t+1}\sim\mathbb P(s_t,a_t)$.
The interaction ends when the agent encounters some terminal state $s_T$. 

The value function  $V^\pi:\mathcal{S}\to\mathbb{R}$ of a policy is defined as $V^\pi(s) = \mathbb{E}_{\pi}[\sum_{t=0}^{T-1} \gamma^t r_{t} | s_0=s]$, where $\mathbb E_\pi$ denotes the expectation of following $\pi$ in the MDP and $T$ is a random variable denoting when a terminal state is reached.
Similarly, the state-action value function $Q^\pi:\mathcal{S}\times\mathcal{A}\to\mathbb{R}$ is defined as $Q^\pi(s,a)=\mathbb{E}_{\pi}[\sum_{t=0}^{T-1} \gamma^t r_{t} | s_0=s,a_0=a]$. The advantage $A^\pi$ is then given by $A^\pi(s,a)=Q^\pi(s,a)-V^\pi(s)$. 
We denote $\rho_\pi$ as the distribution over trajectories $\tau=(s_0,a_0,r_0,\dots,s_T)$ sampled by $\pi$ when interacting with the MDP.

During learning, $\pi$ is typically parameterized (e.g., by a neural network), and in this case, we
use $\pi_\theta$ to denote this parameterized policy with learning parameters given by $\theta$. 
The policy gradient theorem \cite{10.5555/3009657.3009806} states that, in order to optimize the RL objective $\mathbb E_{s_0\sim\rho_0}[V^\pi(s_0)]$, a parameterized policy should be updated with respect to the loss function,
\begin{equation}
    J_{\text{PG}}(\pi_\theta) =\mathbb{E}_{\tau \sim \rho_{\pi_\theta}}\left[\sum_{t=0}^{T-1} \gamma^t \cdot \hat Q_t \log \pi_{\theta}(a_t|s_t) \right],
\end{equation}
 where $\hat Q_t$ is an unbiased estimate of $Q^\pi(s_t,a_t)$. In the simplest case, $\hat Q_t$ is the empirically observed future discounted return following $s_t,a_t$. In other cases, an approximate $Q$-value or advantage function is incorporated to trade-off between the bias and variance in the policy gradients.
 When the $Q$ or $V$ value function is parameterized, we will use $\psi$ to denote its parameters. For example, the policy gradient loss with a parameterized $Q_\psi$ is given by,
 \begin{equation}
    \label{eq:pg}
    J_{\text{PG}}(\pi_\theta, Q_\psi) =\mathbb{E}_{\tau \sim \rho_{\pi_\theta}}\left[\sum_{t=0}^{T-1} \gamma^t \cdot Q_\psi(s_t,a_t) \log \pi_{\theta}(a_t|s_t) \right].
\end{equation}
The value function $Q_\psi$ is typically learned via some regression-based temporal differencing method. For example,
 \begin{equation}
    \label{eq:td}
    J_{\text{TD}}(Q_\psi) =\mathbb{E}_{\tau \sim \rho_{\pi_\theta}}\left[\sum_{t=0}^{T-1} (\hat{Q}_t - Q_\psi(s_t,a_t))^2 \right].
\end{equation}

\paragraph{Stochastic Latent Variable Models} 
In our derivations, we will utilize parameterized policies and value functions conditioned on auxiliary inputs given by stochastic latent variables.
That is, we consider a latent space $\mathcal{Z}$, typically a real-valued vector space. We define a parameterized policy that is conditioned on this latent variable as $\pi_{\theta}(a|s,z)$ for $a\in\mathcal{A},s\in\mathcal{S},z\in\mathcal{Z}$; i.e., $\pi_\theta$ takes in states and latent variables and produces a distribution over actions. In this way, one can consider the latent variable $z$ as modulating the behavior of $\pi_\theta$ in the MDP.
During interactions with the MDP or during training, the latent variables themselves are generated by some stochastic process, thus determining the behavior of $\pi_\theta$. For example, in the simplest case $z$ may be sampled from a latent prior $p_{\upsilon^{\text{(Z)}}}(z|s)$, parameterized by $\upsilon^{\text{(Z)}}$. 
Thus, during interactions with the MDP actions $a_t$ are sampled as $a_t\sim\pi_{\theta}(s_t,z_t), z_t\sim p_{\upsilon^{\text{(Z)}}}(s_t)$.
%We learn the parameters of this distribution using a neural network and denote them by $\theta^{\text{(Z)}}$. %We can factorize the predictive latent variable policy distribution as, $\pi_{\theta}(a_t | s_t, z_t) = \int p_{\theta}(a_t | s_t, z_t) p_{\theta^{\text{(Z)}}}(z_t|s_t) dz,$ where we have an action \textit{decoder} $ p_{\theta}(a_t | s_t, z_t)$ which defines our policy and a latent prior $p_{\theta^{\text{(Z)}}}(z_t|s_t)$ over latent variable $z_t$. 

We treat parameterized latent variable value functions analogously. Specifically, we consider a latent space $\mathcal{U}$ and a parameterized $Q$-value function as $Q_{\psi}(s,a,u)$. A prior distribution over these latent variables may be denoted by $p_{\upsilon^{\text{(U)}}}(u|s)$. 

\section{Policy Gradients Incorporating the Future}

%In our method, we utilize the policy gradient objective with the value function loss.  We write this policy loss function with the state-action value function given a single trajectory as,
%\begin{equation}
%    J_{\text{PG}}(\tau; \pi_{\theta}, Q_{\psi}) =  \sum_{t=0}^T \mathbb{E}_{\tau}[ \log \pi_{\theta}(a_t | s_t) Q_{\psi^{}}(s_t,a_t)],
%\end{equation}
%where $\pi:\mathcal{S}\to\Delta(\mathcal{A})$ and $Q:\mathcal{S} \times \mathcal{A} \to \mathbb{R}$.  

Our method aims to allow a policy during training to leverage \textit{future} information for learning control. 
We propose to utilize stochastic latent variable models to this end. 
Namely, we propose to train $\pi_\theta$ and $Q_\psi$ with latent variables $(\mathbf{z},\mathbf{u})=\{(z_t,u_t)\}_{t=0}^{T-1}$ sampled from a learned function $q_\phi(\tau)$ which has access to the full trajectory.
For example, the PGIF form of the policy gradient objective in (\ref{eq:pg}) may be expressed as
\begin{equation}
    \label{eq:pgif-pg}
    J_{\text{PGIF, PG}}(\pi_\theta,Q_\psi, q_\phi) = \mathbb{E}_{\tau\sim\rho_\pi,(\mathbf{z},\mathbf{u})\sim q_\phi(\tau)}\left[\sum_{t=0}^{T-1} \gamma^t\cdot Q_\psi(s_t,a_t,u_t) \log \pi_{\theta}(a_t | s_t,z_t)\right].
\end{equation}
The PGIF form of the temporal difference objective in (\ref{eq:td}) may be expressed analogously as,
 \begin{equation}
    \label{eq:pgif-td}
    J_{\text{PGIF,TD}}(Q_\psi, q_\phi) =\mathbb{E}_{\tau \sim \rho_{\pi_\theta}, \mathbf{u} \sim q_\phi(\tau)}\left[\sum_{t=0}^{T-1} (\hat{Q}_t - Q_\psi(s_t,a_t,u_t))^2 \right].
\end{equation}
It is clear that any RL objective which trains policies and/or value functions on trajectories can be adapted to a PGIF form in a straightforward manner.
For example, in our experiments we will apply PGIF to an LSTM-based PPO~\cite{schulman2017proximal}, SAC~\cite{pmlr-v80-haarnoja18b}, and BRAC~\cite{wu2020behavior}.

While the PGIF-style objectives above adequately achieve our aim of allowing a policy to leverage future trajectory information during training, they also present a challenge during inference. 
When performing online interactions with the environment, one cannot evaluate $q_\phi(\tau)$, since the full trajectory $\tau$ is not yet observed.

Therefore, while we want to give $\pi_\theta$ and $Q_\psi$ the ability to look at the full $\tau$ during training, we do not want their predictions to overly rely on this privileged information.
To this end, we introduce a regularization on $q_\phi$ in terms of a KL divergence from a prior distribution $p_{\upsilon^{\text{}}}(\tau) := \{p_{\upsilon^{\text{}}}(z_t,u_t|s_t)\}_{t=0}^{T-1}$ which conditions $(z_t,u_t)$ only on $s_t$. Thus, in the case of policy gradient, the full loss is,
\begin{equation}
    \label{eq:pgif-kl}
    J_{\text{PGIF-KL}}(\pi_\theta,Q_\psi,q_\phi)=J_{\text{PGIF, PG}}(\pi_{\theta}, Q_{\psi},q_\phi) + \beta \mathbb{E}_{\tau\sim\rho_\pi}\left[D_{\text{KL}}(q_\phi(\tau)\| p_{\upsilon^\text{}}(\tau))\right],
\end{equation}
where $\beta$ is the weight of the divergence term. 
The introduction of this prior thus solves two problems: (1) it encourages the learned policies and value functions to not overly rely on information beyond the immediate state; (2) it provides a mechanism for inference, namely using latent samples from $p_{\upsilon^\text{}}(s)$ when interacting with the environment.

\begin{figure}[]%
\begin{center}
        \subfigure{\includegraphics[height=4.9cm]{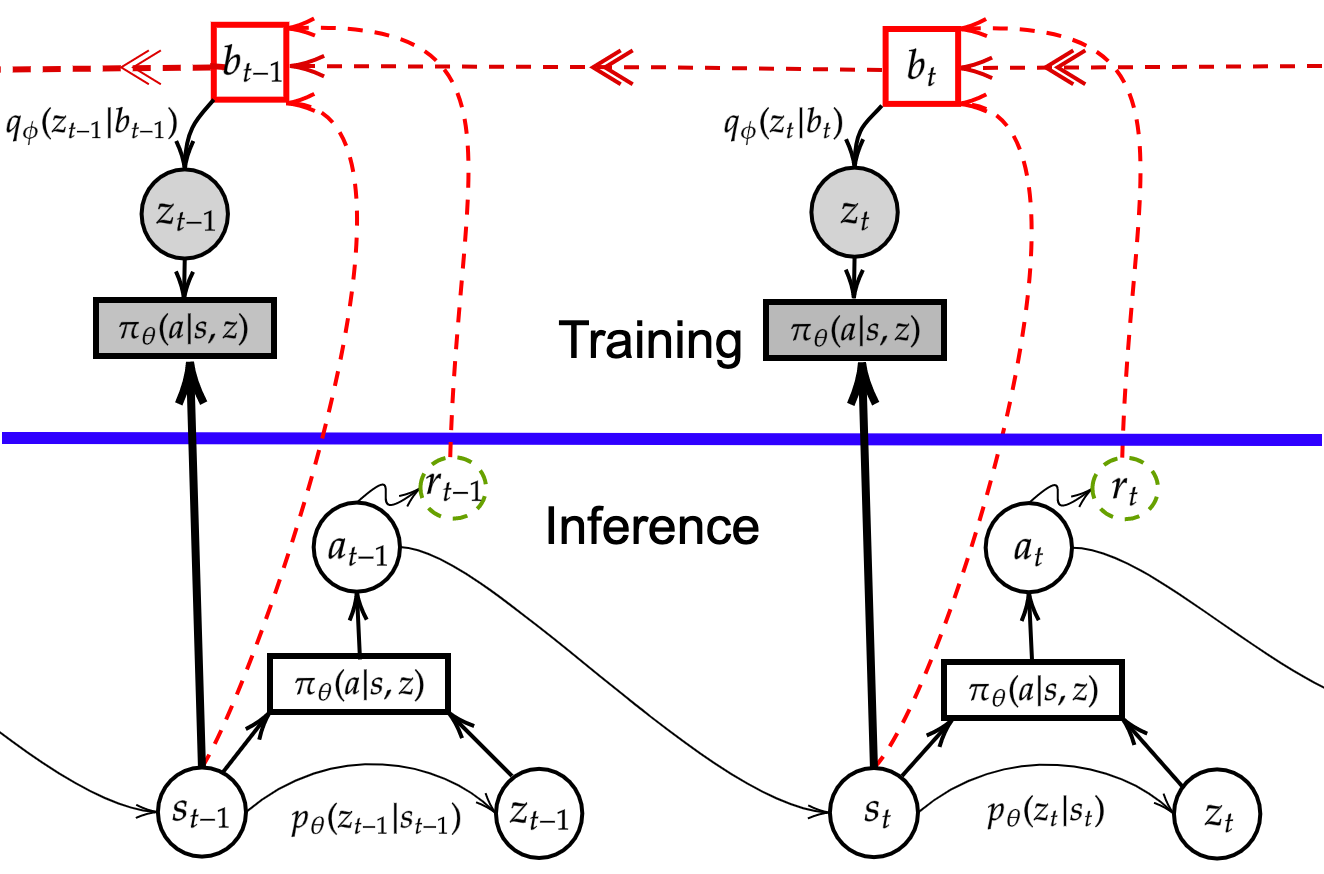}}%
        \end{center}
        \caption{The architecture of the model.  Our inference model $q_{\phi^{(\text{Z})}}$ uses a backwards hidden state $b_t$ to approximate dependencies of $z_t$ on the future of the trajectory.  The blue line separates the data collection and policy gradient training steps in our algorithm and the red lines represent information flowing into the backwards RNN.  Grey variables are used during training and white variables are used during data collection.  \textbf{Top:} we show the training model where the policy gradient loss is calculated with backwards RNN hidden state information. \textbf{Bottom:} we show the data collection phase of the algorithm utilizing latent variables sampled from the latent prior. }%\ofir{Would be great if we included the rest of the model here too, namely the prior and how it is fed into the policy, with some boxes demarcating the difference between what happens in experience collection vs. training}}
        \label{graphical_model}
\end{figure}

\paragraph{Parameterization of $q_\phi$} 
In our implementation, we parameterize $q_\phi(\tau)$ as an RNN operating in reverse order on $\tau$. Specifically, we use an LSTM network to process the states in $\tau$ backwards, to yield LSTM hidden states $\mathbf{b}=\{b_t\}_{t=0}^{T-1}$.
The function $q_\phi$ is then given by Gaussian distributions with mean and variance at time $t$ derived from the backwards state $b_t$.
In practice, to avoid potential interfering gradients from the objectives of $\pi_\theta$ and $Q_\psi$, we use separate RNNs with independent parameters $q_{\phi^{(\text{Z})}}, q_{\phi^{(\text{U})}}$ for $z_t,u_t$, respectively. %For simplicity, we define the joint distribution as $q_{\phi}(z_t,u_t|b_t)$. 
See Figure~\ref{graphical_model} for a graphical diagram of the training and inference procedure for PGIF.  
In our empirical studies below, we will also show that a transformer~\cite{vaswani2017attention} can be used in place of an RNN, providing more computational efficiency and potentially allowing for better propagation of information over time.

\subsection{Variational Information Bottleneck Interpretation}
%\textbf{For this section it may be easier to read the appendix first, since it likely needs some modification due to changes in the derivations.}
The KL regularization we employ above may be interpreted as a variational information bottleneck, constraining the mutual information between the latent variable distribution and the trajectory $\tau$.  
Here we provide a brief derivation establishing this equivalence.

For simplicity, we consider a specific timestep $t\in\mathbb{N}$ and a starting state $s_t=s$. Let $\mathcal{T}_{\geq t}$ denote the random variable for all information contained after and including timestep $t$ in trajectory $\tau$ induced by $\pi$.  Let $\mathcal{U}_t$ be the random variable for latents $u_t$ induced by $q_\phi(\tau_{\geq t})$ conditioned on all steps in the trajectory after and including $t$.  %We define a prior distribution as $p (u_t | s_t=s)$ as a distribution over the latent variable $u_t$ conditioned on the state information variable $s_t \sim \mathcal{S}_t$ at timestep $t$. 
Consider a constrained objective minimizing $J_{\text{PGIF,TD}}$ while enforcing an upper bound $I_{\text{max}}$ on the mutual information between the distribution of trajectory steps and the distribution of latent variables $I(\mathcal{T}_{\geq t},\mathcal{U}_t|s_t=s)$. This objective is given by,
\begin{equation}
\begin{aligned}
    \min_{\psi,\phi} \hspace{2mm} &    J_{\text{PGIF,TD}}(Q_\psi, q_\phi| s_t=s) := \mathbb{E}_{\tau\sim\rho_{\pi_\theta}(\cdot|s_t=s), u_t\sim q_\phi(\tau_{\geq t})}[(\hat{Q}_t - Q_\psi(s_t,a_t,u_t))^2]   \\ 
    &\text{s.t. } I(\mathcal{T}_{\geq t},\mathcal{U}_t|s_t=s) \leq I_{\text{max}}.
    \label{boundImax}
\end{aligned}
\end{equation}
Recall the definition of mutual information:
\begin{equation}
\begin{aligned}
    I(\mathcal{T}_{\geq t},\mathcal{U}_t|s_t=s) &= \int p(\tau_{\geq t},{u}_t | s_t =s) \log \frac{p(\tau_{\geq t},{u}_t| s_t=s )}{p(\tau_{\geq t}|s_t=s)p^{}({u}_t |s_t=s)}d{u}_t d\tau_{\geq t} \\ &= \int \rho_{\pi_{\theta}}(\tau_{\geq t}|s_t=s) q_\phi({u}_t|\tau_{\geq t}) \log \frac{q_\phi({u}_t|\tau_{\geq t})}{p^{}({u}_t | s_t=s)} d{u}_t d\tau_{\geq t}, 
\end{aligned}
\end{equation}
where $p({u}_t|s_t=s)$ is the marginal distribution over the latent variable $p({u}_t|s_t=s) = \int q_\phi({u}_t|\tau_{\geq t}) \rho_\pi(\tau_{\geq t}|s_t=s) d\tau_{\geq t}$. This marginal is intractable to compute directly, and so to approximate this marginal we introduce a variational distribution $h({u}_t |s_t=s)$.  By definition we know that $D_{\text{KL}}[p({u_t}|s_t=s) \lVert h({u_t}|s_t=s)] \geq 0$.  We can then see that $\int p({u}_t|s_t=s) \log p({u}_t|s_t=s) d{u}_t \geq \int p({u}_t|s_t=s) \log h({u}_t|s_t=s) d{u}_t$.  We therefore derive the upper bound for use in \eqref{boundImax} as,
\begin{equation}
\begin{aligned}
    I(\mathcal{T}_{\geq t},\mathcal{U}_t| s_t=s) &\leq \int  \rho_\pi(\tau_{\geq t}|s_t=s) q_\phi({u}_t|\tau_{\geq t}) \log \frac{q_\phi({u}_t|\tau_{\geq t})}{h({u}_t|s_t=s)} d{u}_t d\tau_{\geq t} \\& \leq \mathbb{E}_{\tau \sim \rho_{\pi_\theta}(\cdot|s_t=s)}\Big[D_{\text{KL}}(q_\phi({u}|\tau_{\geq t}) \lVert h({u}_t|s_t=s))\Big].
\end{aligned}
\end{equation}
%This gives rise to a probability divergence interpretation of the bound,
%\begin{equation}
%    I(\mathcal{T},\mathcal{U}) \leq \mathbb{E}_{\tau \sim %\rho_{\pi_\theta}}\Big[D_{{KL}}\Big(\mathcal{E}(u | \tau) \big\lVert %r(u)\Big)\Big].
%\end{equation}
%We can then re-write our original objective as,
%\begin{equation}
%\begin{aligned}
%    \min_{\psi}\hspace{3mm}  &  \mathbb{E}_{\tau \sim %\rho_{\pi_\theta}}\Big[  \mathbb{E}_{u_t,u_{t+1} \sim %\mathcal{E}(u|\tau)}\Big[ J_{\text{PGIF-TD}}(\tau;Q_{\psi}(\cdot, u)) %\Big]\Big] \\ 
%    &\text{s.t. } \mathbb{E}_{\tau \sim %\rho_{\pi_\theta}}\Big[D_{{KL}}\Big(\mathcal{E}(u | \tau) \big \lVert %r(u)\Big)\Big] \leq I_{\text{max}}.
%\end{aligned}
%\end{equation}
We can subsume the constraint into the objective as,
\begin{equation*}
    \begin{aligned}
    \min_{\psi,\phi} \hspace{2mm}&  J_{\text{PGIF,TD}}(Q_\psi, q_\phi|s_t=s) + \beta \left(\mathbb{E}_{\tau \sim \rho_{\pi_{\theta}}(\cdot|s_t=s)} \left[D_{{\text{KL}}}\left(q_\phi(
    \tau_{\geq t})\| h(u_t | s_t=s)\right)\right] -I_{\text{max}}\right).
    \end{aligned}
\end{equation*}
By taking $h$ to be our learned prior $p_\upsilon$, we achieve the single step ($s_t=s$), TD analogue of the PGIF objective in (\ref{eq:pgif-kl}), offset by a constant $\beta\cdot I_\text{max}$.
\subsection{Z-Forcing with Auxiliary Losses}
While our proposed training architecture enables the policy and value function to look at the full trajectory $\tau$, in practice it may be difficult for the trajectory information to propagate, especially in settings with highly stochastic learning signals.  In fact, it is known that such latent variable models may ignore the latent variables due to optimization issues, completely negating any potential benefit \cite{10.5555/2969239.2969370}. 
To circumvent these issues, we make use of the idea of \emph{Z-forcing} \cite{goyal2020recurrent}, which employs auxiliary losses and models to force the latent variables to encode information about the future. 
We denote this loss as $J_{\text{Ax}}(\zeta)$ where $\zeta$ is the set of parameters in any auxiliary models, and elaborate on the main forms of this loss which we consider below.  
%We use the terms $J_{\text{Ax-A}}(\zeta_A)$ and $J_{\text{Ax-C}}(\zeta_C)$ for actor and critic respectively.  Writing our new policy gradient loss with the actor auxiliary yields,
%\begin{equation}
%    J_{\text{PG}}(\theta,\phi,\zeta)= J_{\text{PG-KL}}(\theta,\psi) +\alpha J_{\text{Ax-A}}(\zeta_A)  , 
%\end{equation}
%where $\alpha$ is the weight of the auxiliary loss.

%\subsection{Incorporating reward and transition dynamics}
%We will now discuss some forms of auxiliary loss to exploit different ways of incorporating future dynamics. 

\paragraph{State based forcing (Force)} A simple way to force state information to be encoded is to derive conditional generative models $p_{\zeta}(b_t|z_t)$ over the backwards states given the inferred latent variables $z_t \sim q_{\phi^{{(\text{Z})}}}(z_t | b_t)$, and similarly for the latents $u_t$. We can write this auxiliary objective as a maximum log-likelihood loss $J_{\text{Ax}}(\zeta)=-\mathbb{E}_{q_{\phi^{(\text{Z})}}(z_t | b_t)}[\log p_{\zeta}(b_t | z_t)]$.
This way, we enforce the noisy mapping $b_t\to z_t$ defined by $q_{\phi^{\text{(Z)}}}$ to not be \emph{too noisy} so as to completely remove any information from $b_t$.

\paragraph{Value prediction networks (VPN)} 
A more sophisticated approach to force information to be propagated is to use an autoencoder-like, model-based auxiliary loss. 
To this end, we take inspiration from VPNs \cite{NIPS2017_ffbd6cbb}, and apply an auxiliary loss that uses $b_t$ to predict future rewards, values, and discounts. 
Note that, in principle, $b_t$ already has access to this information, by virtue of the backwards RNN or transformer conditioned on the future trajectory. Thus, this auxiliary loss only serves to enforce that the RNN or transformer dutifully propagates this information from its inputs.
We also note that, in contrast to the state based forcing described above, this approach only enforces $b_t$ to contain the relevant information, and it is up to the RL loss whether this information should be propagated to the latents $z_t,u_t$.

In more detail, the VPN loss introduces the following additional learnable functions:
\begin{tasks}(2)
    \task Encoding network :$f^{(\text{enc})}: s \rightarrow x$ %with abstract state $x \in \mathcal{X}$
    \task Outcome encoder: $f^{(\text{out})}: x, a \rightarrow \hat{\gamma}, \hat{r}$ 
    
    \task Value predictor: $f^{(\text{val})}:x \rightarrow \hat{V}(x)$
    \task Transition predictor: $f^{(\text{trans})}: x,a \rightarrow \hat{x}'$
\end{tasks}
Given a backwards state $b_t$, we first embed this to $x_t^0 = f^{\text{emb}}(b_t)$ with (a). We then make predictions on future rewards $\hat{r}_t^l$, transition dynamics $x_t^l$, discounts $\hat{\gamma}_t^l$, and values $\hat{V}(x_t^l)$ using (b, c, d) for $l=0,\dots,k$.
Finally, we compute the VPN loss as,
\begin{equation}
    J_{\text{VPN}}^{t} = \sum_{l=0}^{k-1} (R_{t+l} - \hat {V}(x_t^l))^2 + (r_{t+l} - \hat{r}_t^l)^2  + (\log_\gamma \gamma_{t+l} - \log_\gamma \hat \gamma_t^l)^2, 
    \label{vpnl}
\end{equation}
where $R_{t+l}$ is the empirically observed future discounted reward in the trajectory.
The full auxiliary loss $J_{\text{Ax}}(\zeta)$ is then given by summing up the loss in (\ref{vpnl}) over all timesteps in all trajectories. 

\subsection{Full Algorithm}

%In practice, we model each distribution discussed with a Gaussian distribution.  The distribution parameters, $[\mu,\sigma^2]$, are learned using neural networks. We sample from these distributions using the reparameterization trick \cite{10.5555/2969442.2969527}. We also adaptively tune the weights of the KL and auxiliary loss.  
The full learning objective for PGIF is thus composed of three components: First, a latent-variable augmented RL objective, e.g., policy gradient as shown in (\ref{eq:pgif-pg}). Second, a KL regularizer, e.g., as shown in (\ref{eq:pgif-kl}).
Finally, an auxiliary loss, given by either state based forcing (Force) or value prediction networks (VPN).
We present an example pseudocode of a PGIF-style policy gradient with learned value function in Algorithm \ref{mainalgo}.

See the Appendix for further details, including how to adaptively tune the coefficients on the KL and auxiliary loss components as well as more specific pseudocode algorithms for advantage policy gradient and soft actor-critic.

\begin{algorithm*}[]
\caption{PGIF Algorithm with State-Action Value Function Estimation}
 \begin{algorithmic}[1]
\REQUIRE Initial parameters: $\theta,\upsilon^{\text{(U)}},\upsilon^{\text{(Z)}},\phi^{\text{(U)}},\phi^{\text{(Z)}},\psi,\zeta_{\text{PG}},\zeta_{\text{TD}}$, Weights: $\alpha_{\text{PG}},\alpha_{\text{TD}},\beta_{\text{PG}},\beta_{\text{TD}}$
\FOR{$\text{policy-step } k=0,1,2,\dots, N$}
\STATE Collect set of Trajectories $\mathcal{D}=\{\tau_i,\dots\}:$
\REPEAT
\STATE $z_{t} \sim p_{\upsilon^{(\text{Z})}}(z_t)$
\STATE Execute: $\pi_{\theta}(a_t|s_t,z_t)$ and observe $r_t, s_{t+1}$ from environment.
\UNTIL{episode termination}
\FOR{$\text{Trajectory: } \tau_i \in \mathcal{D}_{}$}
\STATE $\mathbf{b^Z}$  $=\text{BackwardsLSTM}^\text{Z}(\tau_i)$ (Operates over the entire trajectory)
\STATE $\mathbf{b^U}=\text{BackwardsLSTM}^\text{U}(\tau_i)$
\STATE $\tau_i = \tau_i \cup \{\mathbf{b^Z},\mathbf{b^U}\}$
\ENDFOR
\STATE $\forall \{\mathbf{s},\mathbf{a},\mathbf{r},\mathbf{b^Z},\mathbf{b^U}\} \in \mathcal{D}$: 
%\STATE $\mathbf{z} \sim q_{\phi^{\mathrn{\text{(Z)}}}}(\mathbf{z}|\mathbf{b})$
%\STATE $\mathbf{u} \sim q_{\phi^{(\text{U})}}(\mathbf{u}|\mathbf{b^C})$
%\STATE $\mathbf{u'} \sim p_{\upsilon^{(\text{U})}}(\mathbf{u'}|\mathbf{s'})$
\STATE Derive $J_{\text{Ax-PG}}(\zeta_{\text{PG}}),J_{\text{Ax-TD}}(\zeta_{\text{TD}})$ according to any auxiliary loss.
\STATE $D_{\text{TD}} = D_{{\text{KL}}}(q_{\phi^{{\text{(U)}}}}(\mathbf{u}|\mathbf{b^U}) \lVert p_{\upsilon^{{(\text{U})}}}(\mathbf{u}|\mathbf{s}))$
\STATE $J_{\text{TD}}= \mathbb{E}_{\tau \in \mathcal{D}}\left[J_{\text{PGIF,TD}}(Q_\psi, q_{\phi^{\text{(U)}}})+ \alpha_{\text{TD}} J_{\text{Ax-TD}}(\zeta_{\text{TD}})+\beta_{\text{TD}}  D_{\text{TD}}\right] $
\STATE $D_{\text{PG}} = D_{{\text{KL}}}(q_{\phi^{\text{(Z)}}}(\mathbf{z}|\mathbf{b^Z}) \lVert p_{\upsilon^{{(\text{Z})}}}(\mathbf{z}|\mathbf{s}))$
\STATE $J_{\text{PG}} =\mathbb{E}_{\tau \in \mathcal{D}}\left[J_{\text{PGIF, PG}}(\pi_\theta,Q_\psi, q_{\phi^{\text{(Z)}}}) + \alpha_{\text{PG}} J_{\text{Ax-PG}}(\zeta_{\text{PG}}) + \beta_{\text{PG}}D_{\text{PG}} \right]$ 
\STATE Update all parameters w.r.t: $J_{\text{PG}}$ and $J_{\text{TD}}$
\ENDFOR \end{algorithmic}
\label{mainalgo}
\end{algorithm*}

\section{Related Work}
We review relevant works in the literature in this section.

\paragraph{Incorporating the future} 
Recent works in model-based RL have considered incorporating the future by way of dynamically leveraging rollouts of various horizon lengths and then using them for policy improvement \cite{NEURIPS2018_f02208a0}
. Z-forcing and stochastic dynamics models have been applied directly to learning environmental models and for behavioral cloning while incorporating the future but not for online or offline continuous control \cite{ke2018modeling}.  Our present work is unique for incorporating Z-forcing and conditioning on the future in the model-free RL setting. A few other methods explore the future in less direct ways. For example, RL Upside down \cite{schmidhuber2020reinforcement}, uses both reward (or desired return) and state to predict actions, turning RL into a supervised learning problem. They incorporate future information by way of using these desired returns along a specified desired horizon as input.
 %, while we do not know of any work that has directly incorporated future information in a model-free RL setting as we have done.  

\paragraph{Hindsight} Hindsight credit assignment introduces the notion of incorporating the future of a trajectory by assigning credit based on the likelihood of an action leading to an outcome in the future \cite{hca}. These methods were extended using a framework similar to ours, leveraging a backwards RNN to incorporate information in hindsight \cite{pmlr-v139-mesnard21a}.  Still, there are a number of differences compared to our own work. First, in these previous works only the value function (rather than both the value and policy functions) is provided access to the future trajectory. Second, there is no KL information bottleneck; rather information is constrained via an action prediction objective. Third, these previous works do not employ any Z-forcing, while it is well-known that learning useful latent variables in the presence of an autoregressive decoder is difficult without Z-forcing \cite{bayer2015learning}; in fact, in our own preliminary experiments we found our algorithm performs significantly worse without any auxiliary losses.  %The backwards recurrent network in our method is also much more flexible and easier given the unlimited variety of auxiliary objectives we can incorporate. 
Finally, our method is more versatile, being applicable to off-policy and offline RL settings rather than purely on-policy RL, as in these previous works.  Nevertheless, it is an interesting avenue for future work to investigate how to combine the best of both approaches, especially with guarantees on variance reduction of policy gradient estimators \cite{pmlr-v139-nota21a} and hierarchical policy learning \cite{wulfmeier2021dataefficient}.

Value driven hindsight modelling (HiMo) \cite{DBLP:conf/nips/GuezVWBKPSH20} proposes a hindsight value function, which is conditioned on future information in the trajectory. In this work, they primarily use the hindsight value function (separate from the agent’s value function) to learn a low-dimensional representation, which is distilled to a non-hindsight representation that is then used by the actual actor and critic. Thus there are a few key differences from our work: (1) The hindsight information is only used for value prediction and not policy prediction; (2) There are no gradients passing from the RL loss to the representation loss (only the separate value prediction loss is used to train the representation), thus this method is arguably less end-to-end than PGIF; (3) The only mechanism for controlling the amount of information in the representation is through its dimension size and look-ahead, while using a KL penalty is more flexible.

\paragraph{Combining model-based and model-free RL}
Our proposed PGIF is a way of enabling a model-free RL algorithm to ``look into the future''. Although we do not learn explicit models of the MDP, our work may nevertheless be considered as combining elements of both model-based and model-free RL. Previous works have also claimed to do this, although in distinct ways.  For example, previous work has proposed learning a low dimensional encoding of the environment and then performing planning in this abstract representation along with model-free RL \cite{Francois-Lavet_Bengio_Precup_Pineau_2019}.  VPNs \cite{NIPS2017_ffbd6cbb} improve representations in model-based RL, and use a single network along with supervised learning to learn transition and reward dynamics in an abstract representation of the state space.  Attention augment agents utilize dynamics model roll-outs as input during policy optimization \cite{NIPS2017_9e82757e}. Our work takes a very different approach and incorporates a representation of future information using the hidden states of a backwards RNN, while leveraging VPN-style losses only to enforce information in the RNN to be propagated from inputs to outputs.
 %A large selection of works utilize recurrent neural networks with stochastic dynamics  but they have not widely been applied to RL other than learning environment models \citep{ke2018modeling}. Some works have focused on learning deterministic dynamics models in Atari games \citep{chiappa2017recurrent} using RNNs trained with one-step prediction objectives. 
\paragraph{Learning stochastic latent variable RNNs} Deriving an approximate posterior over stochastic latent variables \cite{bayer2015learning} conditioned on a backwards RNN hidden state has been practically useful in RNNs \cite{goyal2020recurrent} and has been able to overcome being trapped in local minima \cite{Karl2017DeepVB}.  Z-forcing \cite{goyal2020recurrent} is able to learn useful latent variables that capture higher level representations even with a strong auto-regressive decoder, by reconstructing hidden states in a backwards RNN. We use the main methods from Z-forcing to force our variables to learn a useful representation of the trajectory for the agent.  Our method employs an auxiliary signal similar to \cite{Karl2017DeepVB} to design a cost with good convergence properties. Our work is a novel application of these stochastic latent variable RNNs in online and offline policy gradient methods. 

\paragraph{Auxiliary objectives in RL} 
Incorporating auxiliary objectives in RL generally takes the form of intrinsic rewards.  State-space density models have been used to derive intrinsic rewards that incentivize exploration, challenging the agent to find states that generate "surprise" \cite{10.5555/3305890.3305962}.   Other works use the prediction error in an inverse model as a metric for curiosity \cite{10.5555/3305890.3305968}.  The auxiliary objectives we utilize in this work serve the purpose of improving the incorporation of future dynamics information into policy optimization.  

\section{Experiments}
We now provide a wide array of empirical evaluations of our method, PGIF, encompassing tasks with delayed rewards, sparse rewards, online access to the environment, offline access to the environment, and partial observability.
%\begin{enumerate}
%    \item Does our incorporating future information increase sample efficiency and reward in both online and offline RL tasks?
%        \item Is our method still able to perform well with partial observability?
%    \item Does our method show benefits in tasks where credit assignment is critical to success?
%\end{enumerate}

In our online RL experiments, we compare against Soft Actor Critic (SAC) \cite{pmlr-v80-haarnoja18b} and Proximal Policy Optimization (PPO) \cite{schulman2017proximal} with an LSTM layer in the policy network.  The comparison with PPO is particularly important since this method also leverages a form of artificial memory of the past (but not forward-looking like in PGIF).  SAC is a state-of-the-art model-free RL method that employs a policy entropy term in a policy gradient objective and shows optimal performance and stability when compared with other online deep RL benchmarks. We show the hyper-parameters for each experiment in the Appendix. In addition, we explore using our method with a transformer as opposed to an LSTM as the backwards networks. 
\subsection{Credit Assignment}
We first aim to show that our method is highly effective in an environment where credit assignment is the paramount objective. We examine the Umbrella-Length task from BSUITE \cite{osband2020bsuite}, a task involving a long sequential episode where only the first observation (a forecast of rain or shine) and action (whether to take an umbrella) matter, while the rest of the sequence contains random unrelated information.  A reward of $+1$ is given at the end of the episode if the agent chooses correctly to take the umbrella or not depending on the forecast.  This difficult task is used to test the agent's ability to assign credit correctly to the first decision.  We evaluate PGIF-versions of SAC and PPO to vanilla SAC~\cite{pmlr-v80-haarnoja18b} as well as a PPO-LSTM~\cite{schulman2017proximal} baseline. Results are presented in Table \ref{bsuitetabl}.  We see that our method is able to achieve best performance, improving on the baselines by at least 50\%. This suggests that the PGIF agent is able to effeciciently and accurately propogate information about the final reward to the initial timestep, more so than either the one-step backups used in SAC or the multi-step return regressions used in PPO-LSTM can.
\begin{table}[H]
\begin{center}
\begin{tabular}{llll}
\toprule
& Method  & Mean Bsuite-score \\
\midrule
& PPO-LSTM & $0.33 \pm 0.09$ \\ 
& SAC & $0.41 \pm 0.03$  \\
& PGIF-PPO  (VPN) & $0.46 \pm 0.09$\\
& PGIF-PPO (Force) & $0.26 \pm 0.10$\\
& PGIF-SAC (VPN) & $\textbf{0.60} \pm 0.04$\\
& PGIF-SAC (Force) & $\textbf{0.58} \pm 0.08$\\
\end{tabular}
\end{center}
\caption{Performance on the Umbrella-Length environment.  We run our model for $1000$ episodes steps over $5$ random seeds. The BSuite score is calculated in terms of the regret normalized $[\text{random}, \text{optimal}] \rightarrow [0,1]$ (higher is better). The number after $\pm$ is the standard deviation.}
\label{bsuitetabl} 
\end{table}
\subsection{Sparse Rewards}
To test the performance of our method in a sparse reward setting, we utilize the AntMaze environment \cite{pmlr-v78-florensa17a} where we have a simple U-shaped maze with a goal at the end.  The agent receives a reward of $+1$ if it reaches the goal (within an L2 distance of 5) and $0$ elsewhere. The Ant starts at one end of a U-shaped corridor and must navigate to the goal location at the other end. This task is particularly challenging since the reward is extremely sparse.  

We again compare PGIF-versions of SAC and PPO to vanilla SAC and PPO-LSTM. Results are presented in Table~\ref{antable}, where we see that the baselines are unable to make any progress on this challenging task, while PGIF is able to solve the task to a point where it successfully navigates to the goal location in the maze 50\% of the time.
We hypothesize our algorithm has benefits in these environments since as soon as it obtains a reward signal it can adapt quickly and make use of the signal by incorporating it in both policy and value optimization, therefore accelerating learning. 
\begin{table}[H]
\begin{center}
\begin{tabular}{llll}
\toprule
& Method  & Mean Episodic Reward \\
\midrule
& PPO-LSTM & $0.00 \pm 0.00$ \\ 
& SAC & $0.00 \pm 0.00$  \\
& PGIF-PPO (VPN) & $0.20 \pm 0.13$\\
& PGIF-PPO (Force) & $0.30 \pm 0.15$\\\
& PGIF-SAC (VPN) & $\textbf{0.40} \pm 0.16$\\
& PGIF-SAC (Force) & $\textbf{0.50} \pm 0.16$\\\
\end{tabular}
\end{center}
\caption{Mean episodic reward on the AntMaze environment over 10 random seeds trained with $3$ million environment steps.  A sparse reward of $+1$ is obtained at the end of the episode if the agent successfully reaches the goal state within an L2 distance of $5$.  The number after $\pm$ is the standard error.}
\label{antable}
\end{table}
\subsection{Partial Observability}
We now aim to show that our method is not only effective in fully-observed Markovian settings, but also in environments with partial observability. This set of experiments uses the MuJoCo robotics simulator \cite{6386109} suite of continuous control tasks.  These are a set of popular environments used in both online and offline deep RL works \cite{pmlr-v97-fujimoto19a,pmlr-v80-fujimoto18a} and provides an easily comparable benchmark for evaluating algorithm sample efficiency and reward performance.  As in previous work~\cite{yang2021representation}, we introduce an easy modification to these tasks to make the environment partially observable thereby increasing the difficulty: We zero-out a random dimension of the state space at each data collection step.  This helps us replicate a more realistic scenario for a robotic agent where not all of the state space is accessible. 

\begin{figure}[]%
\begin{center}
    \subfigure{\includegraphics[height=3.5cm]{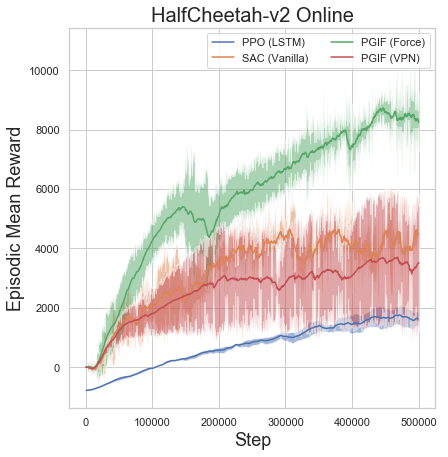}} \hspace{0.0cm}%
        \subfigure{\includegraphics[height=3.5cm]{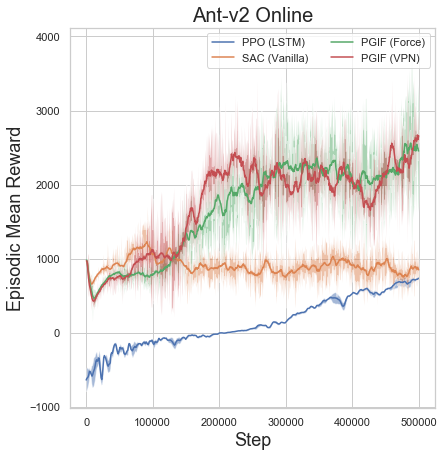}}%
                \subfigure{\includegraphics[height=3.5cm]{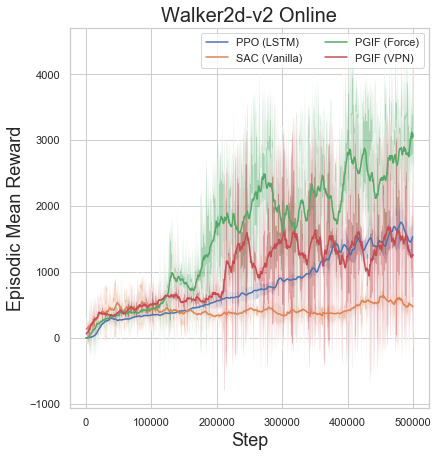}}%
     \subfigure{\includegraphics[height=3.5cm]{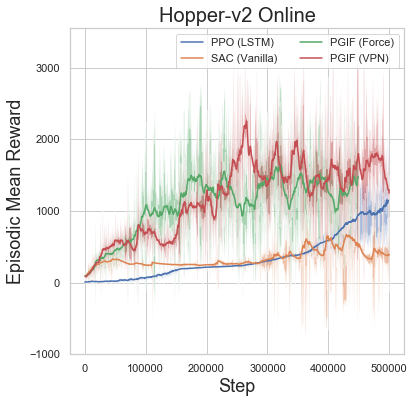}}%
        \end{center}
        
        \caption{The online episodic mean reward evaluated over 5 episodes every 500 steps for MuJoCo continuous control RL tasks with partial observability.  We show the average over $5$ random seeds.  $500,000$ environment step interactions are used. The shaded area shows the standard error.}
        \label{online_exps}
\end{figure}

We compare a PGIF-style SAC implementation to vanilla SAC and PPO-LSTM on these domains.
We show the results of these experiments in Fig. \ref{online_exps}.  
We find that PGIF can provide improved performance on these difficult tasks, suggesting that PGIF is able to leverage future information in the trajectory to appropriately avoid uncertainties about the environment, more so than when only conditioning on the immediate state (vanilla SAC) or even when conditioning on the entire past trajectory (PPO-LSTM).
Interestingly, we find that the simple state based forcing (Force) performs more consistently better than the more sophisticated VPN based forcing.
See the Appendix for additional results, including online evaluations without partial observability. 
\subsection{Offline RL Evaluations}
To asses if our method is effective in an offline RL setting, we evaluate our proposed algorithm in several continuous control offline RL tasks~\cite{fu2020d4rl} against Behavior Regularized Actor Critic (BRAC) \cite{wu2020behavior} and Batch-Constrained Q-learning (BCQ) \cite{pmlr-v97-fujimoto19a}. BRAC operates as a straightforward modification of SAC, penalizing the value function using a measure of divergence (KL) between the behaviour policy and the learned agent policy. For our PGIF algorithm, we use BRAC as the starting point.

For these offline MuJoCo tasks, we examine D4RL datasets classified as \textit{medium} (where the training of the agent is ended after achieving a "medium" level performance) and \textit{medium expert} (where medium and expert data is mixed)~\cite{fu2020d4rl}.  Datasets that contain these sub-optimal trajectories present a realistic problem for offline RL algorithms.  We also include an offline version of the AntMaze, which is particularly challenging due to sparse rewards.  We show the results of these experiments in Table \ref{offtable}. We find that our method outperforms the baselines in all but one of the tasks in terms of final episodic reward.  We hypothesize in the medium-expert setting that we perform slightly worse due to the lack of action diversity which makes learning a dynamics representation difficult.
Interestingly, in contrast to the online results, we find that VPN based forcing performs better than state based forcing, although even state based forcing usually performs better than the baseline methods.
\begin{table}[H]
\begin{center}
\begin{tabular}{llllll}
\toprule
& Environment  & BRAC & PGIF (VPN) & PGIF (Force) & BCQ \\
\midrule
& ant-medium & $2731 \pm 329$  & $\textbf{3250} \pm 125$ & $2980 \pm  164$ & $1851 \pm 94$  \\ 
& ant-medium-expert & $2483 \pm 329$  & $\textbf{3048} \pm 362$  & $2431 \pm  417$ & $2010 \pm 133$ \\
& hopper-medium & $1757 \pm  183$ & $\textbf{2327} \pm 399$  & $1930 \pm 44$ & $1722 \pm 166$ \\
& walker2d-medium & $3687 \pm 25$  & $\textbf{3989} \pm  259$  & $3821 \pm 341$ & $2653 \pm 301$ \\
& halfcheetah-medium & $5462 \pm 198$  & $6037 \pm 324$  & $\textbf{6231} \pm 303$  & $4722 \pm 206$ \\
& halfcheetah-medium-expert & $\textbf{5580} \pm 105$ & $5418 \pm 76$  & $5491 \pm 143$ & $4463 \pm 88$ \\
& antmaze-umaze & $0.5 \pm  0.16$   & $\textbf{0.95} \pm 0.0$   & $0.7 \pm 0.15$ & $0.8 \pm 0.13$  \\
\end{tabular}
\end{center}
\caption{Performance on the offline RL tasks showing the average episodic return. The final average return is shown after training the algorithm for $500,000$ episodes and then evaluating the policy over 5 episodes.  Results show an average of $5$ random seeds. The value after $\pm$ shows the standard error.}
\label{offtable}
\end{table}

\subsection{Transformers Experiments}
\begin{figure}[]%
\begin{center}
    \subfigure{\includegraphics[height=4.5cm]{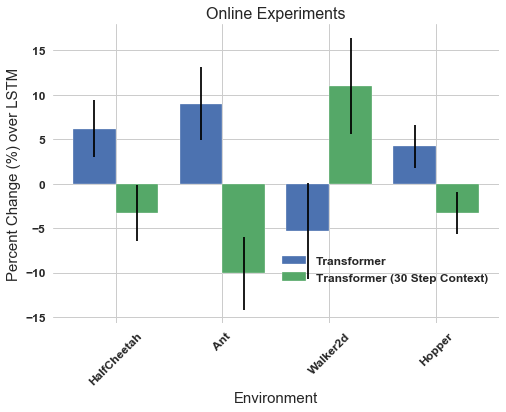}} \hspace{0.2cm}%
        \subfigure{\includegraphics[height=4.5cm]{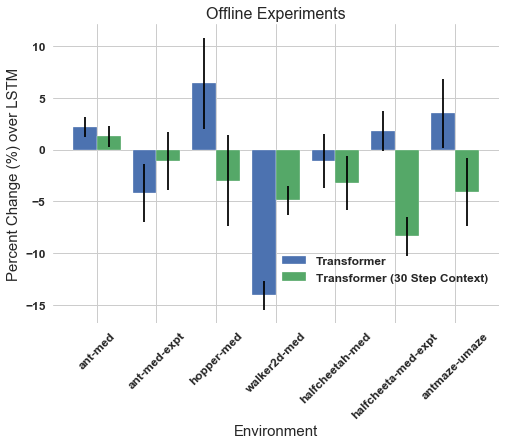}}%
        \end{center}
        \caption{The percent improvement evaluated over 5 episodes by replacing the LSTM backwards network with a transformer.  The $30$ timestep transformer has a context size of a 30 timesteps look-ahead into the future of the trajectory.  We show the average over $5$ random seeds taking the final policy evaluation mean episodic reward as the value.  $500,000$ environment step interactions are used. The error bars shows the standard error.}
        \label{tf}
\end{figure}

Using transformers for processing the future trajectory into latent variables offers a few key benefits over traditional RNN architectures, namely %the ability to process a sequence that is not in order, 
better computational efficiency with long sequences and improved ability to model long timescale interactions.  These provide a few interesting benefits for use with our architecture, namely due to the fact that some of our environments generate sequences up to $1000$ timesteps, which can be onerous for an RNN to process. % and that it is more common to sample batches where the timesteps are not in order. By utilizing a sequence of shuffled observations, batch sampling over multiple different episodes is much simpler.  
In the following experiments, we simply replace our RNN with a transformer. This is inspired by work that models trajectories using a transformer \cite{janner2020sequence} for offline RL.  Additionally, we show the results of experiments where we use a $K$-fixed length context of future states as the input to our transformer.  A short length context is much faster to process than the entire sequence of upstream states, but there is a decrease in performance, suggesting that the entire trajectory contains useful information.  We show these results in Figure \ref{tf}.  The run-time of training was decreased by approximately 40\%.

\section{Discussion}
In this work, we consider the problem of incorporating information from the entire trajectory in model-free online and offline RL algorithms, enabling an agent to use information about the future to accelerate and improve its learning.  %We employ different forms of representation learning to make inferences about future states and rewards and inject them into policy gradients.  
Our empirical results attest to the versatility of our method. The benefits of our method are apparent in both online and offline settings, which is a rare phenomenon given that many previous offline RL works suggest that what works well in online RL often transfers poorly to offline settings, and vice versa.  
Beyond just online and offline RL, our results encompass partial observability, sparse rewards, delayed rewards, and sub-optimal datasets, demonstrating the ability for PGIF to achieve higher reward faster in all settings.

Our work provides many interesting directions for future work.  Namely, one interesting avenue would be to combine our method with exploration strategies in addition to credit assignment.  Perhaps there are benefits in decreasing access to future information during initial exploratory agent steps so the agent is better able to explore. One could also incorporate a metric for surprise to \emph{gate} access to this future information in states of high novelty.  

We also wish to highlight potential risks in our work. Specifically, the use of future information in PGIF may exacerbate biases present in the experience or offline dataset.  For example, it is known that expert datasets lack action diversity.  Further conditioning on the future in this dataset could force these biases to be incorporated more easily into the learning algorithm. Some other biases that may arise are a trajectory that contains some marginal reward and is therefore incorporated into our policy/value function with PGIF. This could hamper exploration of the agent and prevent discovery of states that yield optimal reward. Furthermore, we find that our method is much slower to train than the baselines we compare to, due to the fact that the architecture requires training an LSTM, with at times, very long trajectories as input. We must also use the entire trajectories to train this LSTM rather than transition batches from our replay buffer, further increasing the training complexity. Investigating the transformer architecture rather than an LSTM would be an interesting future direction.

{\small \bibliography{neurips_2021}}
\bibliographystyle{plainnat}

\newpage
\appendix
\section{Implementation Considerations}
We will introduce some implementation details for our model, specifically how the parameters for each distribution are learned. 

\textbf{Data Collection Model:} In the first phase of the actor critic process we are in a \emph{generation} phase of the data. Here, we have a prior over the latent variables $p_{\upsilon^{{\text{Z}}}}(z_t)$.  This prior distribution can be defined as a Gaussian distribution with parameters as $p_{\upsilon^{\text{Z}}}(z_t) = \mathcal{N}(z_t ; \mu^{\text{\text{Z}}}_t, \log \sigma^{\text{Z}}_t)$.  To learn the parameters of this Gaussian prior we utilize an MLP ($f$) defined by $[\mu_t^{\text{Z}}, \log \sigma^{\text{\text{Z}}}_t] =f^{\text{Z}}(s_t)$ which takes the current state as input.  

Now that we have the learned parameters, we can sample $z_t \sim p_{\upsilon^{\text{Z}}}(z_t)$ using the reparameterization trick given the learned mean and standard deviation.  We denote this as  $z_t \sim \text{reparameterize}(\mu_t^{\text{\text{Z}}}, \log \sigma^{\text{Z}}_t)$.  At each time-step during the generation phase, we encounter a state from our environment. We find the distribution parameters and then sample $z_t$ as denoted previously.  The latent variable policy is then executed: $a_t \sim \pi_{\theta}(a_t|s_t,z_t)$.  By applying action $a_t$ on the environment we then obtain $s_{t+1}, r_t$, and repeat until termination.  This leaves us with a sequence of states for use in inference.   We denote the full trajectories generated in this phase with, $\tau$, and $\tau_{(s)} = \{s_0,\dots,s_T\}$ being the trajectories only containing states.  Let us define the set of state only trajectories as, $\{\tau_{(s)}^0,\dots,\tau_{(s)}^n\}\in \mathcal{D}_{(s)}$.

\textbf{Training Model:} The 2nd part of our actor critic algorithm where we learn the policy using the data in the inference phase. This is where we approximate the true posterior distribution over our latent variables. As in Z-forcing, we will use a \emph{backwards} RNN, which takes the reversed sequence of states, $\tau_{(s)}^{\leftarrow}$ as input.  We define this network as,
\begin{equation}
    b_t = f^{\leftarrow}(s_{t+1},b_{t+1}).
\end{equation}
Each state $b_t$ therefore contains information about future states and can be used to shape the approximate posterior distribution over latent variables to contain this information.  We define the inference network with a normal distribution,
\begin{equation}
    q_{\phi^{\text{Z}}}(z_t | b_t) = \mathcal{N}(z_t ; \mu_t^{\text{\text{q}}}, \log \sigma_t^{\text{q}}),
\end{equation}
where, as before, the parameters are learned with an MLP defined by $[\mu_t^{{\text{q}}}, \log \sigma_t^{\text{q}}]=f^{\text{\text{q}}}(b_t)$.

We derive a conditional generative model $p_{\zeta}(b|z)$ over the backwards states given the inferred latent variables $z_t \sim q_{\phi^{\text{\text{Z}}}}(z_t | b_t)$.  Similar to before, we define the parameters of this Gaussian as $p_{\zeta}(b_t|z_t) = \mathcal{N}(z_t; \mu_t^{\text{Ax}}, \log \sigma_t^{\text{Ax}})$, with a MLP that produces $[\mu_t^{\text{Ax}}, \log \sigma_t^{\text{Ax}}] = f^{\text{Ax}}(z_t)$.  %We derive a log likelihood \emph{auxiliary} loss to enforce the latent variables to encode information about the future information encoded in $b_t$ as,
%\begin{equation}
%     \max_{\zeta} \mathbb{E}_{q_{\phi^{\text{U}}}(u | b)}[\log %p_{\zeta}(b | u)].
%\end{equation}

\section{Full Algorithm}
We show the full algorithms for PGIF-PPO (Algorithm \ref{algoppo}) with State based forcing and PGIF-SAC (Algorithm \ref{algosac}) with state based forcing in this section.  Using VPN forcing simply changes the auxiliary loss calculations.  The shared Backwards RNN Pass algorithm is in Algorithm \ref{algbwd}.

\begin{algorithm*}[]
\caption{Algorithm with Advantage Policy Gradient and State Based Forcing}
 \begin{algorithmic}[1]
 \REQUIRE Initial policy parameters: $\theta^{{}}$, Prior parameters: $\upsilon^{{\text{Z}}}, \phi^{{\text{Z}}}$, KL weight: $\beta$, Auxiliary loss Parameters: $\theta^{\text{Ax}}$, Auxiliary loss weight: $\alpha$ Backwards net input parameters: $\theta^{\text{in}}$, Backwards RNN parameters: $\theta^{\text{bkw}}$, Backwards net output parameters:  $\theta^{{\text{bkw-out}}}$, Initial hidden states $\mathbf{h_i}$.
\FOR{$\text{policy-step } k=0,1,2,\dots, N$}
\STATE Collect set of Trajectories $\mathcal{D}=\{\tau_i,\dots\}:$
\REPEAT
\STATE $[\mu_t^{\text{Z}}, \log \sigma^{\text{Z}}_t] =f_{\upsilon^{\text{Z}}}(s_t)$
\STATE $z_t \sim \text{reparameterization}(\mu_t^{\text{Z}}, \log \sigma^{\text{Z}}_t)$
\STATE Execute: $\pi_{\theta^{{}}}(a_t|s_t,z_t) = f_{\theta^{{}}}(s_t,z_t)$
\STATE Observe $s_{t+1}, r_t$ from environment.
\STATE $\tau_i = \tau_i \cup \{s_t,a_t,s_{t+1},r_t\}$
\UNTIL{episode termination}
\STATE Compute advantage estimates $\hat{A}_t^{k}$ using GAE.
\FOR{$\text{Trajectory: } \tau_i \in \mathcal{D}_{}$}
\STATE $\mathbf{b_i}$, $\mathbf{h_{i}}$  $=\text{BackwardsPass}(\tau_i,\mathbf{h}_{{i}}, \theta^{(in)}, \theta^{\text{bkw}},  \theta^{{\text{bkw-out}}})$
\STATE $\tau_i = \tau_i \cup \{\mathbf{b_i}\}$
\ENDFOR
\STATE Reset: $J_{\text{PG}}=0$
\FOR{timestep $t \in \{0,\dots,T\}$}
\STATE $\forall \{s_t,b_t,a_t\} \in \mathcal{D}$: $[\mu_t^{\text{Z}}, \log \sigma_t^{\text{Z-PG}}]=f_{\phi^{\text{Z}}}(b_t,s_t)$
\STATE $z_t^{\text{PG}} \sim \text{reparameterization}(\mu_t^{\text{Z-PG}}, \log \sigma^{{\text{Z-PG}}}_t)$
\STATE $\pi_{\theta^{{}}}(a_t|s_t,z_t^{\text{PG}}) = f_{\theta^{{}}}(s_t,z_t^{\text{PG}})$
\STATE $[\mu_t^{\text{Ax}}, \log \sigma_t^{\text{Ax}}] = f_{\theta^{\text{Ax}}}(z_t^{\text{PG}})$
\STATE $J_{\text{KL}}= \text{KLDivergence}(\mu_t^{\text{Z-PG}}, \log \sigma^{{\text{Z-PG}}}_t, \mu_t^{\text{Z}}, \log \sigma^{\text{Z}}_t)$
\STATE $J_{\text{Ax}} =\text{LogProbGaussian}(b_t, \mu_t^{\text{Ax}}, \log \sigma_t^{\text{Ax}})$
\STATE $J_{\text{PG}} \stackrel{+}{=} \mathbb{E}_{\tau_i \in \mathcal{D}}\Big[ \log \pi_{\theta^{{}}}(a_t|s_t,z_t^{\text{PG}}) \hat{A}^k_t\Big] - \alpha J_{\text{Ax}} - \beta J_{\text{KL}}$ 
\ENDFOR
\STATE Update all parameters w.r.t: $J_{\text{PG}}$
\STATE Update value function estimates with any method.
\ENDFOR
\end{algorithmic}
\label{algoppo}
\end{algorithm*}

\begin{algorithm*}[]
\caption{{BackwardsPass}}
 \begin{algorithmic}[1]
 \REQUIRE Trajectory: $\tau$, Hidden states $h$, Input MLP Parameters: $\theta^{\text{in}}$, Backwards RNN Parameters: $\theta^{\text{bkw}}$, Output net parameters: $\theta^{{\text{bkw-out}}}$.
 \STATE $\tau^{-1}=\text{Reverse-Order}(\tau)$
 \STATE $x_{\text{bkw}} = f_{\theta^{\text{in}}}(\tau^{-1})$
 \STATE $h_{\text{bkw}}=f_{\theta^{\text{bkw}}}(x_{\text{bkw}},h)$ \text{(Operates over the entire sequence)}
 \STATE $x_{\text{bkw-out}} = f_{\theta^{\text{bkw-out}}}(h_{\text{bkw}})$,
 \STATE $x_{\text{bkw-out}}= \text{Reverse-Order}(x_{{\text{bkw-out}}})$ 
  \STATE $h_{\text{bkw}}= \text{Reverse-Order}(h_{{\text{bkw}}}$) 
 \STATE $\textbf{return } x_{\text{bkw-out}}, h_{\text{bkw}}$
 \end{algorithmic}
 \label{algbwd}

\end{algorithm*}

\begin{algorithm*}[]
\caption{SAC Algorithm with State Based Forcing}
 \begin{algorithmic}[1]
 \REQUIRE Initial policy parameters: $\theta^{{}}$, Z forcing parameters: $\upsilon^{{\text{Z}}}, \upsilon^{\text{U}}, \phi^{{\text{Z}}},\phi^{\text{U}}$, KL weight: $\beta$, Auxiliary loss Parameters: $\theta^{\text{Ax-PG}},\theta^{\text{Ax-TD}}$, Auxiliary loss weight: $\alpha$ Backwards net input parameters: $\theta^{\text{in}}$, Backwards RNN parameters: $\theta^{\text{bkw-Z}}, \theta^{\text{bkw-U}} $, Backwards net output parameters:  $\theta^{\text{bkw-out-Z}},\theta^{\text{bkw-out-U}}$, Initial hidden states $\mathbf{h_i}^{\text{Z}},\mathbf{h_i}^{\text{U}}$.
\FOR{$\text{policy-step } k=0,1,2,\dots, N$}
\STATE Collect set of Trajectories $\mathcal{D}=\{\tau_i,\dots\}:$
\REPEAT
\STATE $[\mu_t^{\text{Z}}, \log \sigma^{\text{Z}}_t] =f_{\upsilon^{\text{Z}}}(s_t)$
\STATE $z_t \sim \text{reparameterization}(\mu_t^{\text{Z}}, \log \sigma^{\text{Z}}_t)$
\STATE Execute: $\pi_{\theta}(a_t|s_t,z_t) = f_{\theta}(s_t,z_t)$
\STATE Observe $s_{t+1}, r_t$ from environment.
\STATE $\tau_i = \tau_i \cup \{s_t,a_t,s_{t+1},r_t\}$
\UNTIL{episode termination}
\FOR{$\text{Trajectory: } \tau_i \in \mathcal{D}_{}$}
\STATE $\mathbf{b_i}^{\text{Z}}$, $\mathbf{h_{i}}^{\text{Z}}$  $=\text{BackwardsPass}^{\text{Z}}(\tau_i,\mathbf{h}_{{i}}^{\text{Z}}, \theta^{\text{in-Z}}, \theta^{\text{bkw-Z}},  \theta^{\text{bkw-out-Z}})$
\STATE $\mathbf{b_i}^{\text{U}}$, $\mathbf{h_{i}}^{\text{U}}$  $=\text{BackwardsPass}^{\text{U}}(\tau_i,\mathbf{h}_{{i}}^{\text{U}}, \theta^{\text{in-U}}, \theta^{\text{bkw-U}},  \theta^{\text{bkw-out-U}})$
\STATE $\tau_i = \tau_i \cup \{\mathbf{b_i}^{\text{U}},\mathbf{b_i}^{\text{Z}}\}$
\ENDFOR
\STATE Reset: $J_{\text{PG}},J_{\text{Q}}=0$
\FOR{timestep $t \in \{0,\dots,T\}$}
\STATE $\forall \{s_t,b_t,a_t,s_{t+1},b^{\text{Z}}_t,b^{\text{U}}_t\} \in \mathcal{D}$
\STATE \textbf{Compute $a_{t+1}$}
\STATE $[\mu_{t+1}^{\text{Z}}, \log \sigma_{t+1}^{\text{Z}}]=f_{\upsilon^{\text{Z}}}(s_{t+1})$
\STATE $z_{t+1} \sim \text{reparameterization}(\mu_{t+1}^{\text{Z}}, \log \sigma^{{\text{Z}}}_{t+1})$
\STATE $a_{t+1} \sim \pi_{\theta}(a_{t+1}|s_{t+1},z_{t+1}^{}) = f_{\theta}(s_{t+1},z_{t+1}^{})$
\STATE \textbf{Compute target Q (U-Tar to denote distribution parameters)}
\STATE $[\mu_{t+1}^{\text{U-Tar}}, \log \sigma_{t+1}^{\text{U-Tar}}]=f_{\upsilon^{\text{U}}}(s_{t+1})$
\STATE $u_{t+1}^{\text{Tar}} \sim \text{reparameterization}(\mu_{t+1}^{\text{U-Tar}}, \log \sigma^{\text{U-Tar}}_{t+1})$
\STATE $y(r_t,s_{t+1},d) = r_t + \gamma(1-d)\Big(Q_{\psi\text{-target}}(s_{t+1},a_{t+1},u_{t+1}^{\text{Tar}}) - \alpha \log \pi_{\theta}(a_{t+1}|s_{t+1},z_{t+1}^{}))\Big)$
\STATE \textbf{Update Q functions}
\STATE $[\mu_{t}^{\text{U}}, \log \sigma_{t}^{\text{U}}]=f_{\phi^{\text{U}}}(b_{t}^{\text{Q}},s_t)$
\STATE $u_{t}^{} \sim \text{reparameterization}(\mu_{t+1}^{\text{U}}, \log \sigma^{{\text{U}}}_{t+1})$
\STATE $J_{\text{KL-TD}} = \text{KLDivergence}(\mu_t^{\text{U}}, \log \sigma^{\text{U}}_t, \mu_t^{\text{U-Tar}}, \log \sigma^{\text{\text{U-Tar}}}_t)$
\STATE $[\mu_t^{\text{Ax-TD}}, \log \sigma_t^{\text{Ax-TD}}] = f_{\theta^{\text{Ax-TD}}}(u_t^{\text{}})$
\STATE $J_{\text{Ax-TD}} = \text{LogProbGaussian}(b_t^{\text{Q}},\mu_t^{\text{Ax-TD}}, \log \sigma^{\text{Ax-TD}}_t)$
\STATE $J_Q  \stackrel{+}{=} (Q_{\psi}(s_t,a_t,u_t) -y(r_t,s_{t+1},d)) + \alpha J_{\text{Ax-TD}} + \beta J_{\text{KL-TD}}$
\STATE \textbf{Update Target Networks}
\STATE $(\psi-\text{target}) \leftarrow \rho (\psi-\text{target}) + (1-\rho) \psi$
\STATE \textbf{Update Policy}
\STATE $[\mu_{t}^{\text{Z-PG}}, \log \sigma_{t}^{\text{Z-PG}}]=f_{\phi^{\text{Z}}}(b_{t}^{\text{Z}},s_t)$
\STATE $z_{t}^{\text{PG}} \sim \text{reparameterization}(\mu_{t}^{\text{Z-PG}}, \log \sigma^{{\text{Z-PG}}}_{t})$
\STATE $a_{t} \sim \pi_{\theta}(a_{t}|s_{t},z_{t}^{\text{PG}}) = f_{\theta^{}}(s_{t},z_{t}^{\text{PG}})$
\STATE $[\mu_t^{\text{Ax-PG}}, \log \sigma_t^{\text{Ax-PG}}] = f_{\theta^{\text{Ax-PG}}}(z_t^{\text{PG}})$
\STATE $J_{\text{KL-PG}}= \text{KLDivergence}(\mu_t^{\text{Z-PG}}, \log \sigma^{{\text{Z-PG}}}_t, \mu_t^{\text{Z}}, \log \sigma^{\text{Z}}_t)$
\STATE $J_{\text{Ax-PG}} = \text{LogProbGaussian}(b_t^{\text{Z}}, \mu_t^{\text{Ax-PG}}, \log \sigma_t^{\text{Ax-PG}})$
\STATE $J_{\text{PG}} \stackrel{+}{=} \mathbb{E}_{\tau_i \in \mathcal{D}}\Big[Q_{\psi}(s_t,a_t,u_t^{\text{}}) - \alpha \log \pi_{\theta}(a_t|s_t,z_t^{\text{PG}}) \Big] + \alpha J_{\text{Ax-PG}} + \beta J_{\text{KL-PG}}$ 
\ENDFOR
\STATE Update all parameters w.r.t: $J_{\text{PG}}$ or   $J_{\text{Q}}$
\STATE Update value function estimates with any method.
\ENDFOR
\end{algorithmic}
\label{algosac}

\end{algorithm*}
\pagebreak
\section{Experimental Parameters}
In this section, we give the hyperparameters used for each of our experiments in Tables \ref{hyp1}, \ref{hyp2}, and \ref{trf}.  

\begin{table}[H]
\begin{center}
\begin{tabular}{llll}
\toprule
& Parameter  & Value \\
\midrule
& Optimizer & Adam\\ 
& Learning rate & $5e^{-4}$  \\
& Batch size & $250$\\
& Actor and Critic network dimensions & $(300, 200)$\\\
& RNN Dim & 15 \\
& RNN-Embedding Dim & 20 \\
& Z or Q Dim & 5 \\
& Z-force Neural net dim & 20\\\
& Initial random exploration steps & 10000\\\
& Replay Buffer Size & 1,000,000 steps\\\
& Discount & 0.99\\\
& Evaluation Episodes & 5\\\

\end{tabular}
\end{center}
\caption{Parameters used for PGIF and SAC in the online experiments with MuJoCo.}
\label{hyp1}
\end{table}

\begin{table}[H]
\begin{center}
\begin{tabular}{llll}
\toprule
& Parameter  & Value \\
\midrule
& Optimizer & Adam\\ 
& Learning rate & $1e^{-3}$  \\
& Value penalty & True \\
& Batch size & $250$\\
& Actor and Critic network dimensions & $(300, 200)$\\\
& Q value ensemble & $2$ \\
& RNN Dim & $15$ \\
& RNN-Embedding Dim & $20$ \\
& Z or Q Dim & $10$ \\
& Z-force Neural net dim & $20$\\\
& Initial random exploration steps & $10000$\\\
& Replay Buffer Size & $1,000,000$ steps\\\
& Discount & $0.99$\\\
& Evaluation Episodes & $5$\\\
& BRAC Value Penalty $\alpha$ & $0.1$ \\

\end{tabular}
\end{center}
\caption{Parameters used for transformers experiments}
\label{hyp2}
\end{table}

\begin{table}[H]
\begin{center}
\begin{tabular}{llll}
\toprule
& Parameter  & Value \\
\midrule
& Hidden Dim & 128 \\
& Activation & ReLu \\
& Heads & 1 \\
& Heads & 1 \\
& Layers & 3 \\
& Attention Dropout & 0.1 \\
\end{tabular}
\end{center}
\caption{Parameters used for the transformers experiments.}
\label{trf}
\end{table}

Training the behavior policy for BRAC offline RL experiments uses behavior cloning with 300,000 time steps.  

For the purpose of the backwards RNN, since we have variable episode length we pad the episode sequence to the maximum length with zeros.  The maximum number of timesteps is generally $1000$ in MuJoCo.  The training of the RNN with this long sequence and the pre-processing steps involved are expensive computationally.  For offline experiments, our replay buffer is episodic, so we train with an entire episode at each update.  For online we train with a batch of randomly sampled steps from the replay buffer (and obtain the respective episodes that those steps were in for the training of the backwards RNN).

For the weight parameters $\alpha$ and $\beta$, we recommend that the weight on the KL be increased at a small fixed constant rate during training so that the final returned policy uses minimal information about the future. We use an initial weight of $1e^{-4}$ for the KL and auxiliary losses and increase it by $1e^{-6}$ at each training iteration for experiments to a max weight of $1$. Thus, at the end of training the policy and value functions should be minimally reliant on this privileged information.

\section{Online Experiment Learning Curves}
\begin{figure}[H]%
\begin{center}
    \subfigure{\includegraphics[height=4cm]{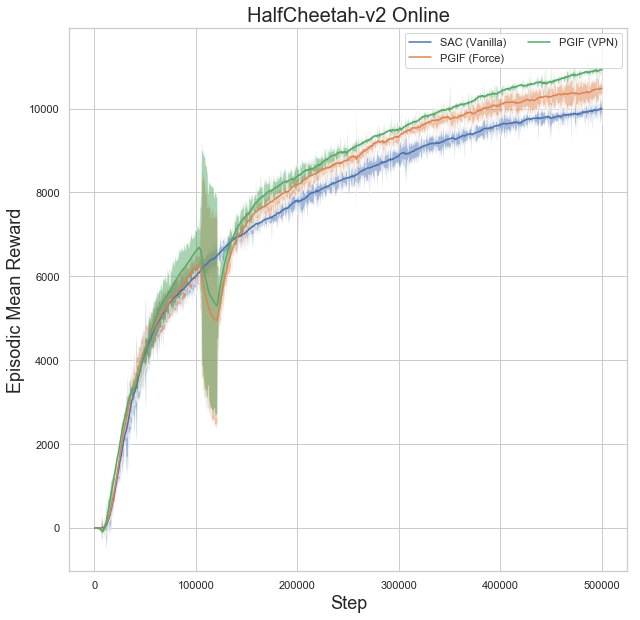}} \hspace{0.0cm}%
        \subfigure{\includegraphics[height=4cm]{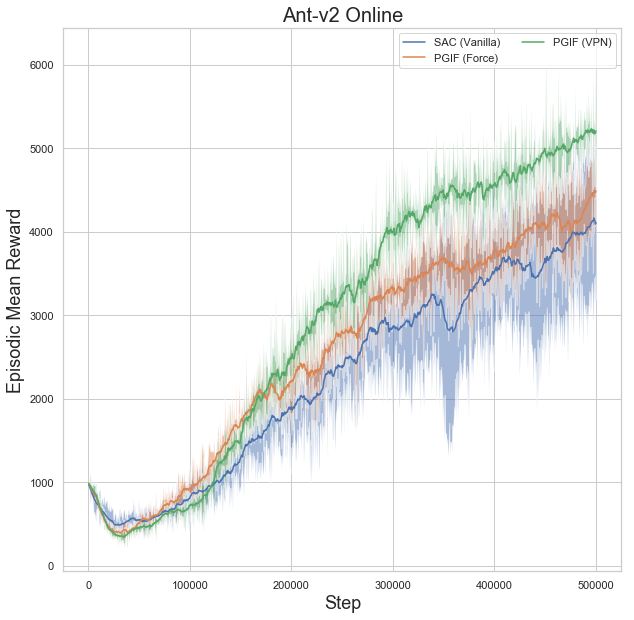}}%
        \hspace{3.0cm}
                \subfigure{\includegraphics[height=4cm]{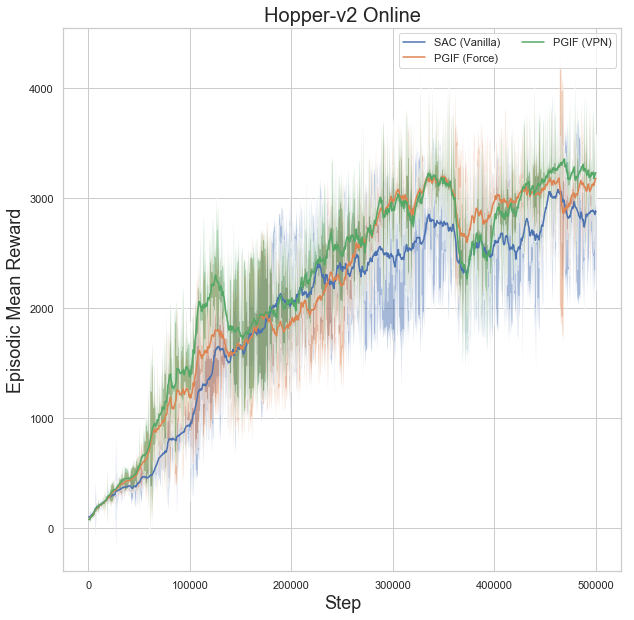}}%
     \subfigure{\includegraphics[height=4cm]{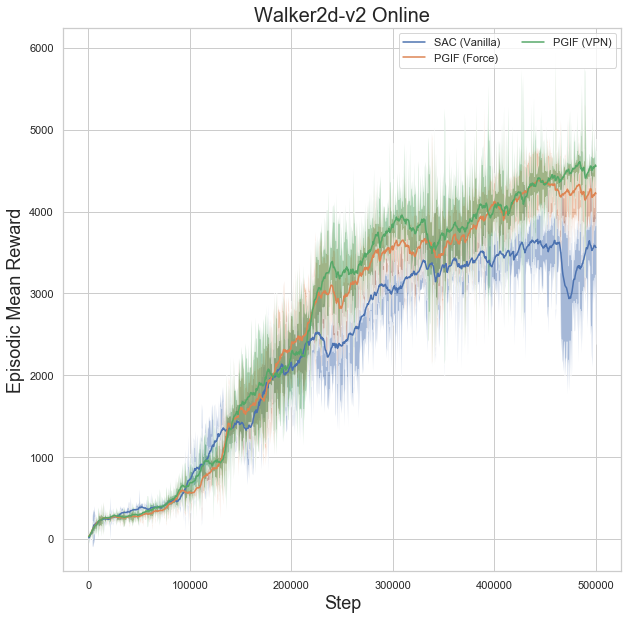}}%
        \end{center}
        
        \caption{The online episodic mean reward evaluated over 5 episodes every 500 steps for MuJoCo continuous control RL tasks with full observability.  We show the average over $5$ random seeds.  $500,000$ environment step interactions are used. The shaded area shows the standard error.}
        \label{online_exps_lc}
\end{figure}

\begin{figure}[H]%
\begin{center}
    \subfigure{\includegraphics[height=3.4cm]{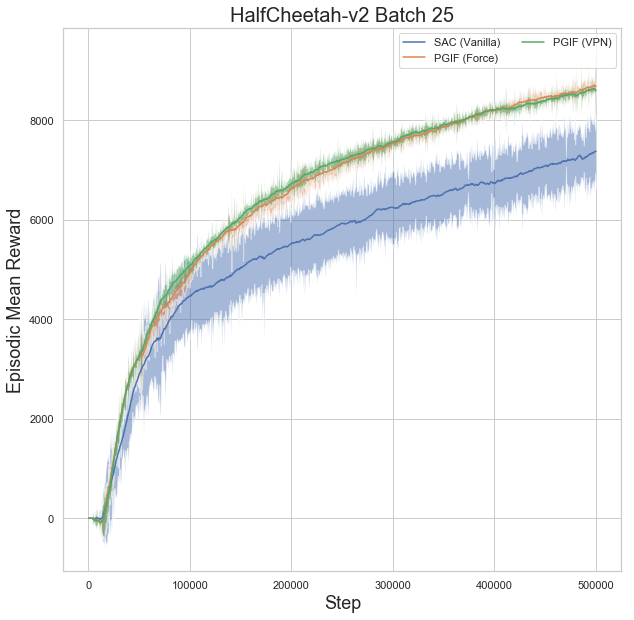}} \hspace{0.0cm}%
        \subfigure{\includegraphics[height=3.4cm]{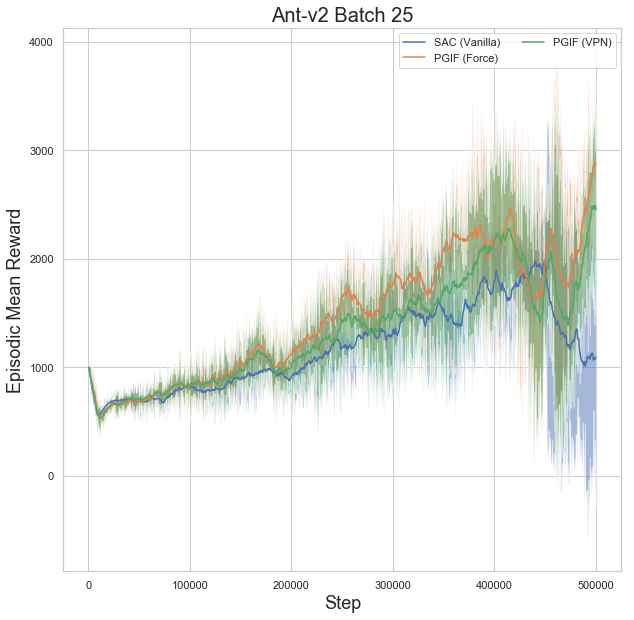}}%
                \subfigure{\includegraphics[height=3.4cm]{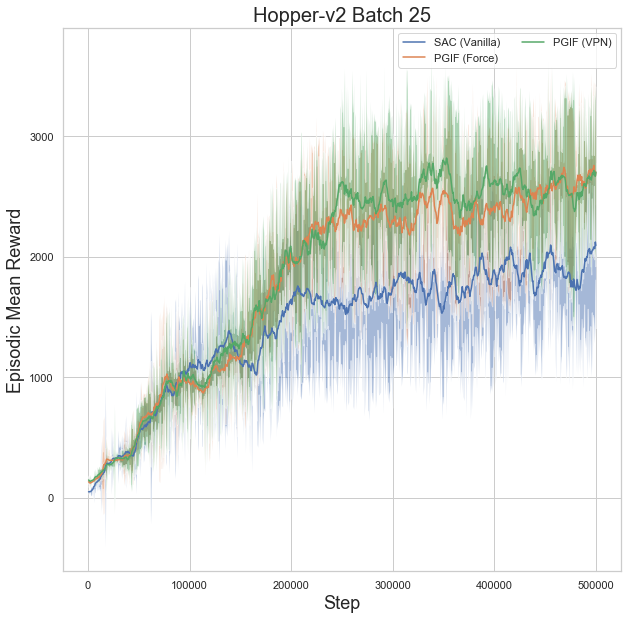}}%
     \subfigure{\includegraphics[height=3.4cm]{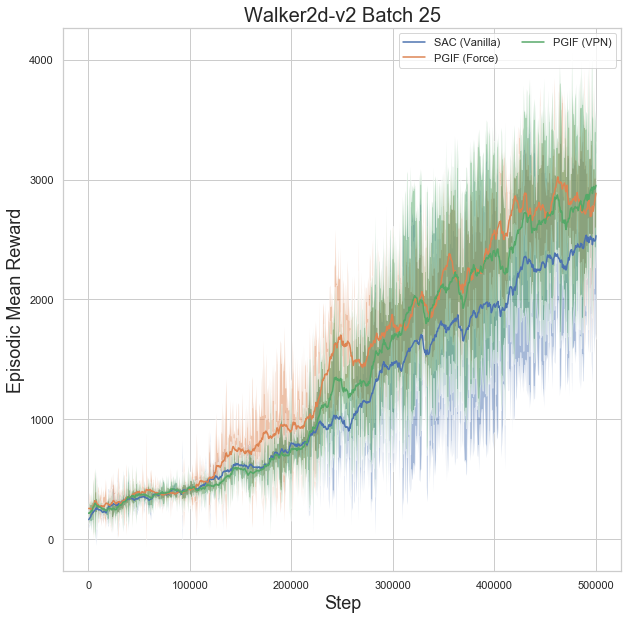}}%
        \end{center}
        
        \caption{The online episodic mean reward evaluated over 5 episodes every 500 steps for MuJoCo tasks with full observability.  We show the average over $5$ random seeds.  $500,000$ environment step interactions are used.  The model is updated with a batch size of 25. The shaded area shows the standard error.}
        \label{online_exps_25}
\end{figure}

In this section, we show the training curves for the online experiments with full observability in Figure \ref{online_exps_lc}.  We compare PGIF against SAC only, since we see that PPO performs much worse from initial experiments.  Note that in practicality, it is best to reduce the maximum number of steps in the MuJoCo environments to 500 for better computational efficiency in the backwards RNN steps (without much change in final reward).  We note that our method appears to have a small reduction in standard error compared to vanilla SAC.
We aim to investigate the effect PGIF learning with a lower batch size.  We notice that PGIF has better performance at lower batch sizes.  It may be an interesting direction to further investigate this anomaly.   In Figure \ref{online_exps_25}, we show the performance of PGIF with a batch size of 25.  

\section{Computing Infrastructure}
The cluster used to run these experiments has $688$ NVIDIA V100-SXM2 GPUs.
\end{document}